\documentclass{article}
\PassOptionsToPackage{numbers,compress}{natbib}
\usepackage[preprint]{neurips_2025}
\usepackage[utf8]{inputenc}
\usepackage[T1]{fontenc}
\usepackage{microtype}
\usepackage{amsmath}
\usepackage{amssymb}
\usepackage{booktabs}
\usepackage{tabularx}
\usepackage{enumitem}
\usepackage{xcolor}
\usepackage{colortbl}
\usepackage{graphicx}
\usepackage{float}
\usepackage{tikz}
\usetikzlibrary{arrows.meta,positioning,fit,backgrounds,calc,shadows,decorations.pathreplacing}
\usepackage{pgfplots}
\pgfplotsset{compat=1.18}
\usepgfplotslibrary{fillbetween}

\definecolor{nAccent}{HTML}{BB480B}
\definecolor{nAccentSoft}{HTML}{FEF3EC}
\definecolor{nInk}{HTML}{222222}
\definecolor{nBody}{HTML}{4B5563}
\definecolor{nMuted}{HTML}{6B5B4A}
\definecolor{nGray}{HTML}{8A8174}
\definecolor{nBorder}{HTML}{D6CFC0}
\definecolor{nBorderLight}{HTML}{EDE6DA}
\definecolor{nPaper}{HTML}{FAF7F1}

\usepackage[font=small,labelfont={bf,color=nAccent},labelsep=period]{caption}
\usepackage[colorlinks=true,linkcolor=nAccent,citecolor=nAccent,urlcolor=nAccent,hyperfootnotes=false]{hyperref}
\usepackage{xurl}
\arrayrulecolor{nBorder}

\newsavebox{\quotebox}

\newcommand{\thead}[1]{\textbf{#1}}
\newcommand{\rowrule}{\arrayrulecolor{nBorderLight}\specialrule{0.4pt}{3pt}{3pt}\arrayrulecolor{nBorder}}
\newcommand{\tablestyle}{}
\newcommand{\solvedbar}[2]{\tikz[baseline=-0.6ex]{%
  \fill[nBorderLight] (0,-0.09) rectangle (1.4,0.09);
  \fill[nInk!70] (0,-0.09) rectangle ({1.4*#1/#2},0.09);}}
\newcommand{\costbar}[1]{\tikz[baseline=-0.6ex]{%
  \fill[nBorderLight] (0,-0.09) rectangle (1.4,0.09);
  \fill[nGray] (0,-0.09) rectangle ({1.4*#1/0.55},0.09);}}

\setlist{itemsep=2pt,topsep=3pt}
\newcommand{\code}[1]{\texttt{\small #1}}
\newcommand{\todo}[1]{\textcolor{red}{[TODO: #1]}}

\newcommand{\fusescreenshot}[1]{%
  \IfFileExists{#1}{%
    \tikz\node[draw=nBorder, line width=0.5pt, inner sep=0pt, rounded corners=1pt]{\includegraphics[width=\linewidth]{#1}};}{%
    \fbox{\parbox[c][5cm][c]{0.95\linewidth}{\centering\todo{save screenshot as \texttt{\detokenize{#1}}}}}}%
}

\title{Numina Fuse: Interactive Autoformalization in Lean~4}
\author{%
  Justin Asher$^{2*\dagger}$\hspace{1.4em}
  Yichuan Cao$^{1*}$\hspace{1.4em}
  Ruichen Qiu$^{1*}$\hspace{1.4em}
  Pietro Pellegrino$^{2,3}$\hspace{1.4em}\\[0.5ex]
  \textbf{Jia Li}$^{2\dagger}$\\[1.6ex]
  \small\normalfont $^{1}$Academy of Mathematics and Systems Science, University of Chinese Academy of Sciences \\[0.3ex]
  \small\normalfont $^{2}$Project Numina \\[0.3ex]
  \small\normalfont $^{3}$University of Naples Federico II
}

\begin{document}
\maketitle
{\renewcommand{\thefootnote}{}\footnotetext{$^{*}$Equal contribution.\quad $^{\dagger}$Project leads.}\footnotetext{Correspondence to Justin Asher (\texttt{justin@projectnumina.ai}).}}

\begin{abstract}
Numina Fuse is an agent-native application for formalizing mathematics in Lean~4. Since Lean can compile while not representing its intended informal meaning, and agents now write Lean faster than people can read it, we built Fuse around human supervision of agent work. A user works with an AI orchestrator on a Lean project and supervises it through views of the blueprint, the dependency graph, and the Lean code, while the backend keeps the project's derived state and runs agents for hours without the user present. We describe the development of Fuse throughout 2026, as it moved from a prescribed pipeline of agents paced by human review to an orchestrator that dispatches specialists, provers, and parallel children on its own. We designed for this shift, keeping project state, tooling, and interface design separate from orchestration so that agents could take over more of the work as models improved. We report Fuse's use in a formalization of the three-dimensional Kakeya conjecture and a study of prover models. We close with lessons for the next generation of autoformalization systems. Fuse Desktop, an open-source local version, is available at \url{https://github.com/project-numina/fuse-desktop}.
\end{abstract}

%=====================================================================
\section{Introduction}

Numina Fuse is an agent-native application for formalizing mathematics in Lean~4. We built it in early 2026, when formalization benefited significantly from mathematicians overseeing the work of AI agents. Fuse was designed so that human experts could review that work in real time, from the structure of the blueprint to the faithfulness and quality of the formalization. Each of the three problems shaped part of the design: reviewer agents and human statement review address faithfulness, an explorer and the project's dependency graph support reuse, and a live view of every agent action, which the user can interrupt and redirect, addresses drift.

The user works with an AI orchestrator on a Lean project while the page shows the blueprint (the informal mathematics, linked to its Lean declarations), the dependency graph, the Lean code with its goals and errors, and each agent action as it happens. The user can redirect the agent mid-task, edit the blueprint or the Lean code directly, and review the result as an ordinary pull request. Behind the interface, the backend builds and indexes every project, caching the dependency graph, build errors, declaration locations, the Mathlib API relevant to each lemma, and text extracted from source papers. Because everything runs on our servers, agents keep working after the user closes the browser.

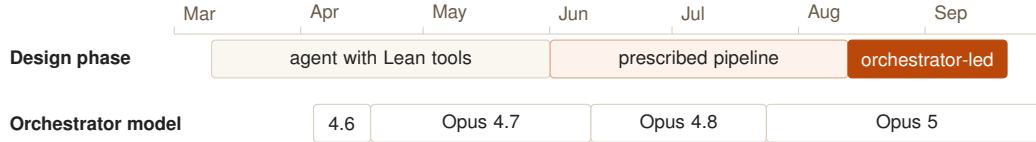
\begin{figure}[t]
\centering
% Timeline of Fuse's design and orchestrator model, March to September 2026.
% Dates from git history and design records. Drawn at true size.
\begin{tikzpicture}[font=\scriptsize\sffamily,
  rowlabel/.style={anchor=west, font=\scriptsize\sffamily\bfseries, text=nInk, align=left},
  band/.style={rounded corners=1.5pt, minimum height=5mm, inner sep=1pt, anchor=west, align=center},
  added/.style={anchor=north west, align=left, text=nInk, inner sep=1.5pt, font=\scriptsize\sffamily}]
\draw[nBorder, line width=0.4pt] (2.300,0.33) -- (2.300,0.43);
\node[text=nMuted, anchor=west] at (2.300,0.6) {Mar};
\draw[nBorder, line width=0.4pt] (3.973,0.33) -- (3.973,0.43);
\node[text=nMuted, anchor=west] at (3.973,0.6) {Apr};
\draw[nBorder, line width=0.4pt] (5.592,0.33) -- (5.592,0.43);
\node[text=nMuted, anchor=west] at (5.592,0.6) {May};
\draw[nBorder, line width=0.4pt] (7.265,0.33) -- (7.265,0.43);
\node[text=nMuted, anchor=west] at (7.265,0.6) {Jun};
\draw[nBorder, line width=0.4pt] (8.885,0.33) -- (8.885,0.43);
\node[text=nMuted, anchor=west] at (8.885,0.6) {Jul};
\draw[nBorder, line width=0.4pt] (10.558,0.33) -- (10.558,0.43);
\node[text=nMuted, anchor=west] at (10.558,0.6) {Aug};
\draw[nBorder, line width=0.4pt] (12.231,0.33) -- (12.231,0.43);
\node[text=nMuted, anchor=west] at (12.231,0.6) {Sep};
\draw[nBorder, line width=0.4pt] (13.850,0.33) -- (13.850,0.43);
\draw[nBorder, line width=0.4pt] (2.3,0.33) -- (13.85,0.33);
\node[rowlabel] at (0,0) {Design phase};
\node[rowlabel] at (0,-0.85) {Orchestrator model};
\node[band, minimum width=4.480cm, text width=4.380cm, fill=nPaper, draw=nBorder, text=nInk] at (2.786,0) {agent with Lean tools};
\node[band, minimum width=3.940cm, text width=3.840cm, fill=nAccentSoft, draw=nAccent!50, text=nInk] at (7.265,0) {prescribed pipeline};
\node[band, minimum width=2.105cm, text width=2.005cm, fill=nAccent, draw=nAccent, text=white] at (11.205,0) {orchestrator-led};
\node[band, minimum width=0.756cm, text width=0.656cm, fill=white, draw=nBorder, text=nInk] at (4.135,-0.85) {4.6};
\node[band, minimum width=2.914cm, text width=2.814cm, fill=white, draw=nBorder, text=nInk] at (4.891,-0.85) {Opus 4.7};
\node[band, minimum width=2.321cm, text width=2.221cm, fill=white, draw=nBorder, text=nInk] at (7.805,-0.85) {Opus 4.8};
\node[band, minimum width=3.724cm, text width=3.624cm, fill=white, draw=nBorder, text=nInk] at (10.126,-0.85) {Opus 5};
\end{tikzpicture}
\caption{Fuse's design phases from March to September 2026, from git history and our internal design records, with the orchestrator's default model.}
\label{fig:timeline}
\end{figure}

Fuse changed substantially while we built it. It began as a single agent with Lean tools, became a prescribed pipeline of writer, reviewer, and prover agents with the human as the final check, and ended as an orchestrator that dispatches long-lived specialists, batches of provers, and parallel delegated orchestrators on its own (Figure~\ref{fig:timeline}).

We built Fuse with the bitter lesson in mind, as we knew hand-built structure helps only until general methods, scaled with better models, overtake it~\cite{sutton2019bitter}. We kept what should outlast any model (project state, Lean tooling, interface design) separate from orchestration, so agents could take over more of the work as models improved.

A Numina team used Fuse's stages and human checkpoints for much of a formalization of the three-dimensional Kakeya conjecture~\cite{guth2026streamlined}. Eventually, there came a crossover point where using swarms of unmonitored agents became more effective than agents guided by users. Consequently, in the final two weeks, one researcher directing Claude Opus~5 and the newly released Claude Fable~5.1 outside Fuse formalized the remainder of the conjecture's proof (Section~\ref{sec:eval}).

This paper describes and evaluates the hosted system. We have since open-sourced Fuse Desktop, a single-user version that shares the hosted app's interface and agent roles and runs locally with the user's own Claude Code or Codex account (Figure~\ref{fig:screenshot}). Desktop runs those roles as subagents of the user's Claude Code session, or within a single Codex session, without the hosted app's prover routing, delegation, sandbox, or metering; the interface figures in this paper are taken from it.

This paper includes three contributions:
\begin{itemize}[leftmargin=1.2em]
\item A description of an agent-native environment for formalization that connects human statement review, agent proof search, persistent project state, and Lean verification, as deployed (Sections~\ref{sec:using}--\ref{sec:infra}).
\item A record of how that design changed over six months of model releases, and the observed failures that drove each change (Sections~\ref{sec:agents} and~\ref{sec:lessons}).
\item Deployment experience on the Kakeya formalization, including Lean output rates for Fuse and for a directed coding agent, and a prover selection and cost study (Section~\ref{sec:eval}).
\end{itemize}

%=====================================================================
\section{Using Fuse}
\label{sec:using}

\begin{figure}[!b]
\centering
\fusescreenshot{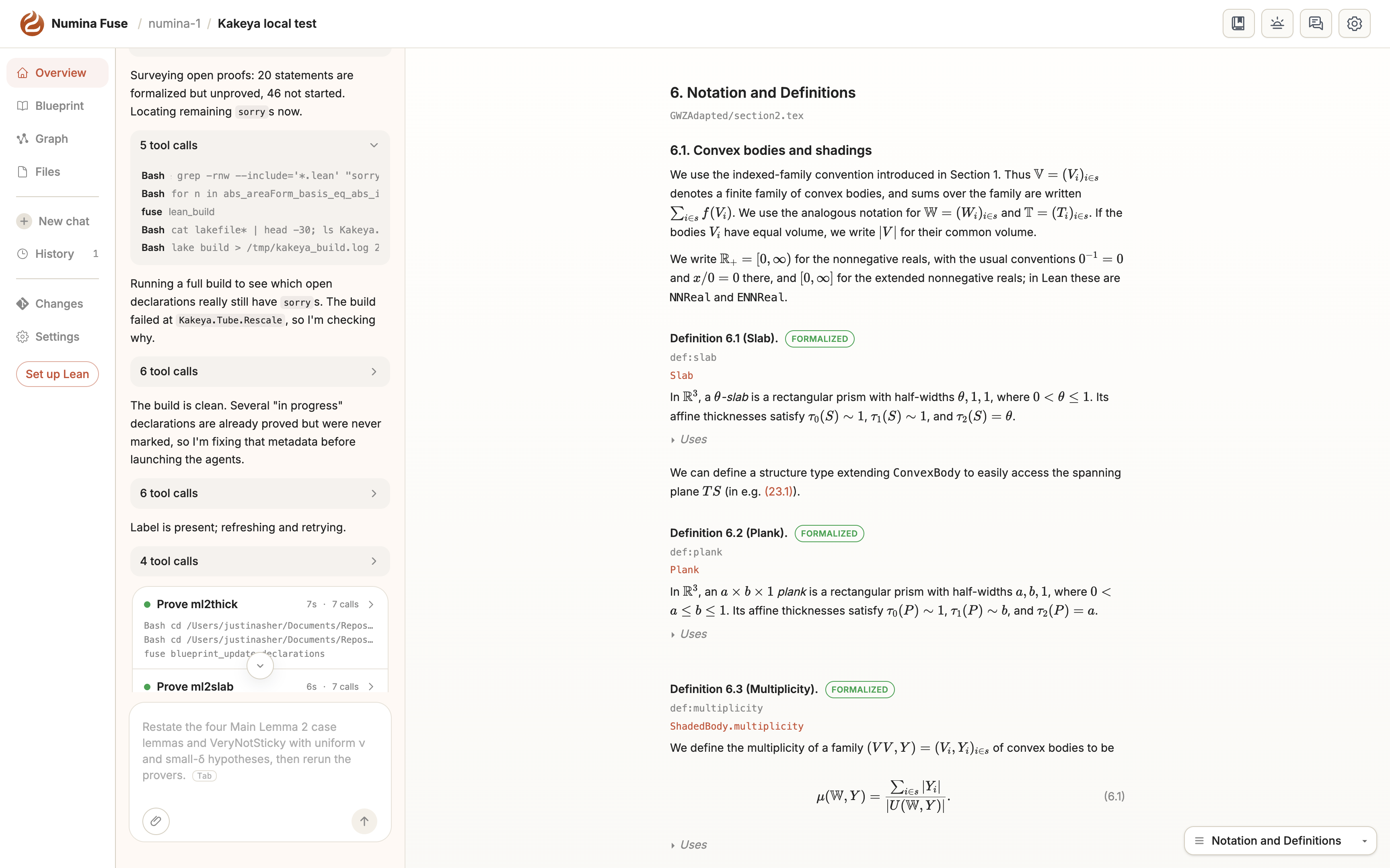}
\caption{Fuse Desktop (\url{https://github.com/project-numina/fuse-desktop}), the open-source local version, on the Kakeya formalization. Left: the chat with the orchestrator, which narrates its plan, groups its tool calls, and shows provers running as Claude Code subagents (\emph{Prove ml2thick}, \emph{Prove ml2slab}). Right: the rendered blueprint, with each definition's label, linked Lean declaration, and formalization status.}
\label{fig:screenshot}
\end{figure}

A project in Fuse is a GitHub repository containing one or more Lean packages. The user opens a \emph{workspace}, a branch of that repository with a blueprint attached, and uploads the papers being formalized. Agents draft a blueprint from the source, or the user adopts an existing one. Each declaration in the blueprint links to the Lean declaration that formalizes it, and its status (not started, formalized, proved) is shown on the dependency graph.
The workspace gathers the project's views in one place: an overview of progress, a viewer for the source papers, the blueprint organized by chapter, a dependency graph for each chapter, and a Lean editor with the goals and diagnostics from the language server. Each blueprint entry shows its informal statement and links to its Lean declaration, with its proof and dependencies in collapsible sections. Appendix~\ref{app:gallery} shows these views.

The user then asks the orchestrator for an outcome (Figure~\ref{fig:screenshot}), such as ``formalize Section~5'' or ``prove the remaining lemmas in this file''. The orchestrator dispatches specialists and provers, each of which appears in the chat as a card streaming its tool calls, edits, and verdicts; edits appear as diffs, and builds report errors as they happen. The user can message the orchestrator agent at any point to steer it. Every session runs in the background, survives the browser closing, and notifies the user when it needs attention.

Fuse allows people to evaluate the quality of the Lean formalizations. In the Kakeya project, the most consequential human interventions were statement corrections. In one case an agent, unable to prove a lemma, replaced its sorry-free statement with a stronger one that was not provable, and a mathematician caught the change in review. Three similar statement errors on the Kakeya branch (an unsatisfiable hypothesis, a Lean statement stronger than its blueprint entry, and a quotient that was false as stated) were all found by people after the fact, and motivated the statement checks described in Section~\ref{sec:agents}. Work reaches the main branch only through a pull request that a person merges.

%=====================================================================
\section{Architecture}
\label{sec:arch}

\subsection{What lasts and what changes}
\label{sec:goals}

Fuse separates agent orchestration from other processes. Work lives in the user's repository as an ordinary Lean and leanblueprint project, so it survives without Fuse and is reviewed as a pull request. Derived state (parsed blueprints, dependency graphs, build diagnostics, agent notes) lives in the backend, so every agent, whatever its model, starts from the same picture of the project. Sessions run on our servers and survive both a closed browser and a deployment. The orchestration layer is the prompts, roles, and dispatch rules that tell agents what to do, which changed with nearly every model release (Section~\ref{sec:agents}).

\subsection{Components}

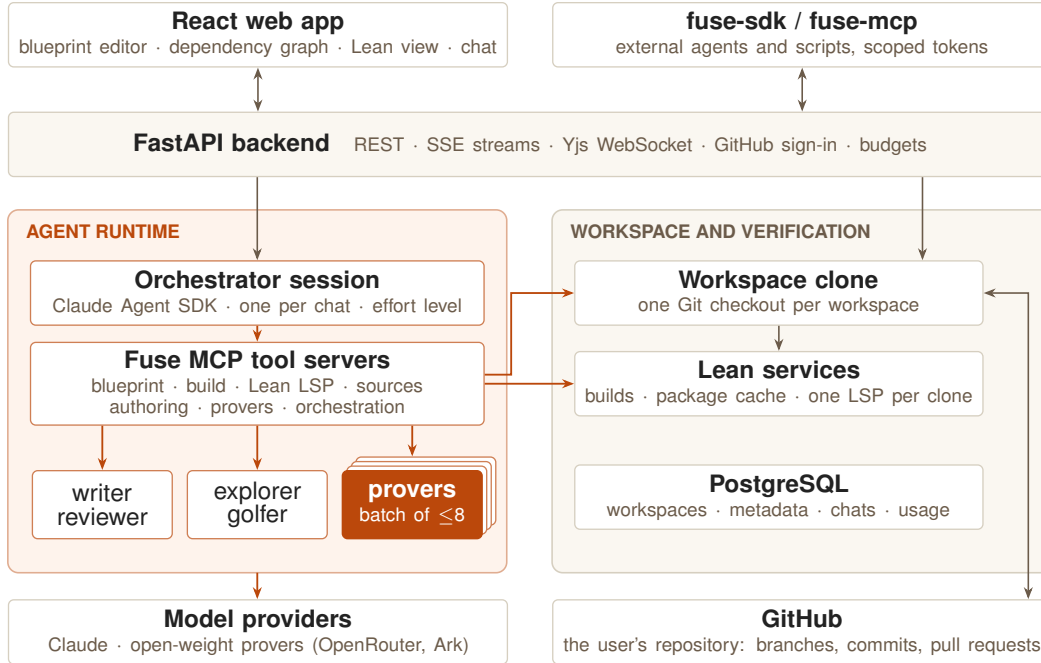
\begin{figure}[t]
\centering
% Fuse system architecture. Drawn at true size for a 5.5in text block.
\providecommand{\figsub}{\scriptsize\sffamily\color{nMuted}}
\begin{tikzpicture}[
  font=\footnotesize\sffamily,
  >={Stealth[length=1.6mm]},
  box/.style={draw=nBorder, fill=white, rounded corners=2pt, line width=0.5pt,
    minimum height=8.5mm, align=center, inner sep=3pt, text=nInk},
  agentbox/.style={box, draw=nAccent!70},
  sub/.style={font=\scriptsize\sffamily, text=nMuted},
  panel/.style={rounded corners=4pt, line width=0.6pt},
  ptitle/.style={font=\scriptsize\sffamily\bfseries, text=#1, anchor=north west},
  flow/.style={->, line width=0.6pt, draw=nMuted},
  aflow/.style={->, line width=0.7pt, draw=nAccent},
  lbl/.style={font=\scriptsize\sffamily\itshape, text=nMuted, fill=white, inner sep=1pt},
]
% ---------------------------------------------------------------- clients
\node[box, minimum width=6.6cm, text width=6.35cm, anchor=north west] (fe) at (0,0)
  {\textbf{React web app}\\[-1pt]{\figsub blueprint editor $\cdot$ dependency graph $\cdot$ Lean view $\cdot$ chat}};
\node[box, minimum width=6.6cm, text width=6.35cm, anchor=north west] (sdk) at (7.2,0)
  {\textbf{fuse-sdk / fuse-mcp}\\[-1pt]{\figsub external agents and scripts, scoped tokens}};

% ---------------------------------------------------------------- backend bar
\node[box, fill=nPaper, minimum width=13.8cm, text width=13.55cm, anchor=north west] (api) at (0,-1.45)
  {\textbf{FastAPI backend}\quad{\figsub REST $\cdot$ SSE streams $\cdot$ Yjs WebSocket $\cdot$ GitHub sign-in $\cdot$ budgets}};
\draw[flow, <->] (fe.south) -- (fe.south |- api.north);
\draw[flow, <->] (sdk.south) -- (sdk.south |- api.north);

% ---------------------------------------------------------------- agent panel
\fill[panel, nAccentSoft] (0,-2.75) rectangle (6.6,-7.55);
\draw[panel, nAccent!45] (0,-2.75) rectangle (6.6,-7.55);
\node[ptitle=nAccent] at (0.12,-2.82) {AGENT RUNTIME};

\node[agentbox, minimum width=6.0cm, text width=5.75cm] (orch) at (3.3,-3.85)
  {\textbf{Orchestrator session}\\[-1pt]{\figsub Claude Agent SDK $\cdot$ one per chat $\cdot$ effort level}};
\node[agentbox, minimum width=6.0cm, text width=5.75cm] (mcp) at (3.3,-5.05)
  {\textbf{Fuse MCP tool servers}\\[-1pt]{\figsub blueprint $\cdot$ build $\cdot$ Lean LSP $\cdot$ sources}\\[-2pt]{\figsub authoring $\cdot$ provers $\cdot$ orchestration}};
% child agents: three chips
\node[agentbox, minimum width=1.85cm, text width=1.60cm, minimum height=9mm] (c1) at (1.25,-6.65) {writer\\[-1pt]reviewer};
\node[agentbox, minimum width=1.85cm, text width=1.60cm, minimum height=9mm] (c2) at (3.3,-6.65) {explorer\\[-1pt]golfer};
% prover stack, drawn as several offset cards to show the batch
\foreach \k in {3,2,1}
  \node[agentbox, minimum width=1.85cm, text width=1.60cm, minimum height=9mm, fill=white]
    at ({5.35+0.06*\k},{-6.65+0.06*\k}) {};
\node[agentbox, minimum width=1.85cm, text width=1.60cm, minimum height=9mm, fill=nAccent, text=white, draw=nAccent] (c3) at (5.35,-6.65)
  {\textbf{provers}\\[-1pt]{\scriptsize batch of $\le$8}};
\draw[aflow] (orch) -- (mcp);
\draw[aflow] (mcp.south -| c1) -- (c1.north);
\draw[aflow] (mcp.south -| c2) -- (c2.north);
\draw[aflow] (mcp.south -| c3) -- ($(c3.north)+(0,0.18)$);
\draw[flow] (api.south -| orch) -- (orch.north);

% ---------------------------------------------------------------- workspace panel
\fill[panel, nPaper] (7.2,-2.75) rectangle (13.8,-7.55);
\draw[panel, nBorder] (7.2,-2.75) rectangle (13.8,-7.55);
\node[ptitle=nMuted] at (7.32,-2.82) {WORKSPACE AND VERIFICATION};

\node[box, minimum width=5.4cm, text width=5.15cm] (clone) at (10.2,-3.85)
  {\textbf{Workspace clone}\\[-1pt]{\figsub one Git checkout per workspace}};
\node[box, minimum width=5.4cm, text width=5.15cm] (lean) at (10.2,-5.05)
  {\textbf{Lean services}\\[-1pt]{\figsub builds $\cdot$ package cache $\cdot$ one LSP per clone}};
\node[box, minimum width=5.4cm, text width=5.15cm] (db) at (10.2,-6.55)
  {\textbf{PostgreSQL}\\[-1pt]{\figsub workspaces $\cdot$ metadata $\cdot$ chats $\cdot$ usage}};
\draw[flow] (clone) -- (lean);
\draw[flow] ($(api.south -| clone)+(1.9,0)$) -- ($(clone.north)+(1.9,0)$);

% agents act on the workspace only through tools
\draw[aflow] (mcp.east) -- (lean.west |- mcp.east);
\draw[aflow] ($(mcp.east)+(0,0.12)$) -- ++(0.35,0) |- (clone.west);

% ---------------------------------------------------------------- externals
\node[box, minimum width=6.6cm, text width=6.35cm, anchor=north west, fill=white] (llm) at (0,-7.9)
  {\textbf{Model providers}\\[-1pt]{\figsub Claude $\cdot$ open-weight provers (OpenRouter, Ark)}};
\node[box, minimum width=6.6cm, text width=6.35cm, anchor=north west, fill=white] (gh) at (7.2,-7.9)
  {\textbf{GitHub}\\[-1pt]{\figsub the user's repository: branches, commits, pull requests}};
\draw[aflow] (3.3,-7.55) -- (3.3,-7.9);
\draw[flow, <->] (clone.east) -- (13.5,-3.85) -- (13.5,-7.9);
\end{tikzpicture}
\caption{Fuse architecture. The orchestrator acts on a workspace through MCP tool servers and a restricted shell, and delegates heavy work to child processes whose events stream back to the backend.}
\label{fig:arch}
\end{figure}

Figure~\ref{fig:arch} shows the main components of Fuse. The user works in a React application whose views (blueprint editor, dependency graph, Lean goals and diagnostics, Git, chat) all read from a FastAPI backend, which streams agent output and workspace events to the browser as they happen. Each session drives an orchestrator, a Claude Code or Codex process working on its own clone of the repository, which acts through Fuse's Model Context Protocol (MCP) tool servers (Table~\ref{tab:tools}) and starts child agents as separate processes. The backend builds each clone with a Lean service and keeps durable state in PostgreSQL, so the browser is only a view.

\begin{table}[b]
\centering\footnotesize\tablestyle
\caption{MCP tool servers available to the orchestrator. Most are started per session and bound to the session's clone.}
\label{tab:tools}
\begin{tabularx}{\linewidth}{@{}lX@{}}
\toprule
\thead{Server} & \thead{Representative tools} \\
\midrule
blueprint & read blueprint and declarations; update entries; set declaration status; refresh metadata \\
\rowrule
build & build project or module; build status, errors, warnings; measure heartbeats \\
\rowrule
lean-lsp & goal state, hover, diagnostics, declaration resolution, Loogle search \\
\rowrule
lean-explore & semantic search over Mathlib and other indexed packages \\
\rowrule
sources & list sources; read text, OCR (lazy, per page), PDF pages as images; search \\
\rowrule
authoring & \code{explore}, \code{draft\_blueprint}, \code{review\_blueprint}, \code{formalize}, \code{golf}, \code{review\_proofs} \\
\rowrule
prover & \code{run\_provers} (batch proof search) \\
\rowrule
orchestration & \code{delegate\_tasks}, \code{auto\_complete}, task cancellation, metadata recovery \\
\bottomrule
\end{tabularx}
\end{table}

\subsection{Project state}

\paragraph{Blueprints and workspaces.} A workspace is a managed branch, identified by the repository and branch, with a blueprint attached; a repository may hold several Lean projects, and each workspace records which one it targets. A parser expands \code{\textbackslash input} files and extracts each declaration's label, linked Lean names, dependencies, and proof block. The data is stored with PostgreSQL, partitioned by owner: parser-owned fields are recomputed on every refresh, agent-owned fields (assessments, notes, Mathlib API found relevant) are written only by agents. Lean declaration locations are resolved by Lean itself rather than by text search. Source papers are stored outside Git and exposed to agents with lazily computed, cached OCR.

\paragraph{Recorded, checked, and reviewed status.}
\label{sec:status}
Fuse tracks three kinds of evidence about a declaration:
\begin{itemize}[leftmargin=1.2em, itemsep=1pt, topsep=2pt]
\item \emph{Recorded status} is what the blueprint's \code{\textbackslash leanok} markers say; any person or agent can set them.
\item \emph{Checked status} is what Lean confirms: after a prover finishes, the backend asks the language server whether the target declarations compile without errors or \code{sorry}, and downgrades the recorded status if not.
\item \emph{Reviewed faithfulness}, whether the Lean statement means what the source says, comes only from reviewer agents and people.
\end{itemize}
These checks are deliberately light. Fuse does not run axiom checks or compare statements against their originals after every turn: as in ordinary development, a project passes through incomplete states, and we expected residual issues to be fixed once most of the formalization was complete. The Kakeya release was verified this way, by a script that rebuilds the library and checks the exact axioms of the main theorem (Section~\ref{sec:eval}).

%=====================================================================
\section{Agents}
\label{sec:agents}

\paragraph{Runtime.} A session is a Claude Code or Codex process driven through its SDK, with Fuse's system prompt, role definitions, MCP servers, and a permission callback. Administrators choose the harness, model, and reasoning effort for each role; by default the orchestrator is Claude Opus and provers use an inexpensive model through an Anthropic-compatible endpoint. Every SDK message becomes a typed event streamed to the browser, and every model invocation is recorded with its model, provider, and token counts.

\paragraph{Sandboxing.} Running coding agents on a shared server against users' repositories was a persistent source of work. Agents must not reach other users' clones, credentials, or the host, and must not damage the repository they work on. A harness's own permission settings were not enough: some built-in tools skip the permission callback, and native subagents do not inherit Fuse's checks, so every child is started by Fuse. A permission gate limits the shell to an allowlist (read-only Git, module-scoped builds, and a few inspection commands, each with its own forbidden flags), confines paths to the clone, and refuses writes to Git internals and CI workflows; only the backend commits and pushes. Because building a Lean file an agent wrote can run arbitrary code during elaboration, the final boundary is an operating-system jail with namespaces and resource limits around each agent process.

\paragraph{Specialists and provers.}
Work is divided among specialized subagents: a blueprint writer and reviewer, a formalizer and formalizer-reviewer, an explorer that finds reusable Mathlib and project API, a golfer that shortens proofs, a proof-style reviewer, and provers.

The explorer searches Mathlib and other indexed packages with lean-explore, queries Loogle for declarations by type, and reads the project's own blueprint declarations, their dependencies, and its Lean code, so that writers and provers reuse existing results instead of restating them.

Writers and reviewers are long-lived specialists that communicate with each other. The reviewer returns a verdict per declaration, ready or needing revision, and sends its findings directly to the writer. After finding errors in our Kakeya formalization, we increased the responsibility the reviewers, with blueprint reviewer first looks for counterexamples and unsatisfiable hypotheses, and the formalizer reviewer compares each Lean statement with its blueprint entry.

Proof search runs as batches of provers, up to eight at a time and one per declaration, each a Claude Code process pointed at an inexpensive model and working in a scratch copy of its file.

\begin{figure}[!b]
\centering
% The authoring pipeline. Drawn at true size for a 5.5in text block.
\providecommand{\figsub}{\scriptsize\sffamily\color{nMuted}}
\begin{tikzpicture}[
  font=\footnotesize\sffamily,
  >={Stealth[length=1.6mm]},
  stage/.style={draw=nBorder, fill=white, rounded corners=2pt, line width=0.5pt,
    minimum width=1.7cm, text width=1.6cm, minimum height=1.3cm, inner sep=1pt, align=center, text=nInk},
  loopstage/.style={stage, draw=nAccent!70, fill=nAccentSoft},
  num/.style={circle, fill=nAccent, text=white, inner sep=0pt, minimum size=3.6mm,
    font=\tiny\sffamily\bfseries},
  flow/.style={->, line width=0.6pt, draw=nMuted},
  art/.style={font=\scriptsize\sffamily\itshape, text=nMuted, align=center},
  brace/.style={decorate, decoration={brace, amplitude=3pt}, draw=nAccent, line width=0.5pt},
]
\def\xs{0.85}\def\dx{2.03}
% stages
\node[stage]      (s1) at ({\xs+0*\dx},0) {\textbf{Source}\\[-1pt]{\figsub PDF, LaTeX,}\\[-3pt]{\figsub or notes}};
\node[stage]      (s2) at ({\xs+1*\dx},0) {\textbf{Explore}\\[-1pt]{\figsub find API}\\[-3pt]{\figsub to reuse}};
\node[loopstage]  (s3) at ({\xs+2*\dx},0) {\textbf{Draft}\\[-1pt]{\figsub writer $\rightleftarrows$}\\[-3pt]{\figsub reviewer}};
\node[loopstage]  (s4) at ({\xs+3*\dx},0) {\textbf{Formalize}\\[-1pt]{\figsub writer $\rightleftarrows$}\\[-3pt]{\figsub reviewer}};
\foreach \k in {3,2,1}
  \node[stage, draw=nAccent!60] at ({\xs+4*\dx+0.06*\k},{0.06*\k}) {};
\node[stage, fill=nAccent, draw=nAccent, text=white] (s5) at ({\xs+4*\dx},0)
  {\textbf{Prove}\\[-1pt]{\scriptsize\sffamily one prover}\\[-3pt]{\scriptsize\sffamily per lemma}};
\node[stage]      (s6) at ({\xs+5*\dx},0) {\textbf{Golf}\\[-1pt]{\figsub shorten,}\\[-3pt]{\figsub recheck}};
\node[stage]      (s7) at ({\xs+6*\dx},0) {\textbf{Style}\\[-1pt]{\figsub remove model}\\[-3pt]{\figsub artifacts}};
% numbers
\foreach \i in {2,...,7} {
  \pgfmathtruncatemacro{\n}{\i-1}
  \node[num] at ($(s\i.north west)+(0.05,-0.05)$) {\n};
}
% arrows and artifacts
\foreach \a/\b in {1/2,2/3,3/4,4/5,5/6,6/7}
  \draw[flow] (s\a.east) -- (s\b.west);
\node[art, below=3pt of s3] {$\le$3 passes\\[-2pt]blueprint \texttt{.tex}};
\node[art, below=3pt of s4] {$\le$3 passes\\[-2pt]Lean, \texttt{sorry}'d};
\node[art, below=3pt of s5] {\phantom{x}\\[-2pt]checked proofs};
\node[art, below=3pt of s7] {\phantom{x}\\[-2pt]pull request};
% loop annotations
% statement gates brace
\draw[brace] ($(s3.north west)+(0,0.12)$) -- node[above=3pt, font=\scriptsize\sffamily, text=nAccent, align=center]
  {statement gates} ($(s4.north east)+(0,0.12)$);
\draw[brace, draw=nGray] ($(s5.north west)+(0,0.12)$) -- node[above=3pt, font=\scriptsize\sffamily, text=nMuted]
  {inexpensive models, $\le$8 parallel} ($(s5.north east)+(0.18,0.12)$);
% failure feedback
\draw[->, line width=0.6pt, draw=nAccent, dashed]
  ($(s5.south)+(-0.3,-0.85)$) to[out=200, in=-20] node[below, font=\scriptsize\sffamily\itshape, text=nAccent]
  {failed: relaunch, or revise the statement} ($(s4.south)+(0.3,-0.85)$);
\end{tikzpicture}
\caption{The authoring stages and their order. Shaded stages are writer/reviewer pairs, whose reviewers carry the statement checks. In the pipeline phase, the orchestrator's instructions and one blocking tool per stage enforced this order; in the final design, each child orchestrator runs these stages in the order its region needs.}
\label{fig:pipeline}
\end{figure}
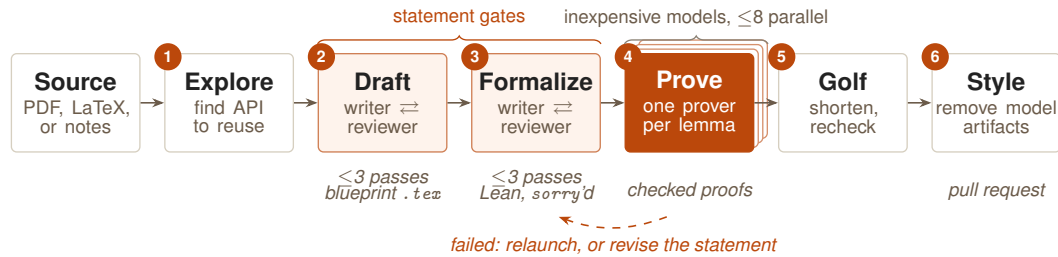

\paragraph{From a pipeline to orchestrator-led dispatch.} In its pipeline phase (Figure~\ref{fig:timeline}), Fuse ran the stages of Figure~\ref{fig:pipeline} in a fixed order, each as a blocking tool call. When the orchestrator took over dispatch, we bounded it with rules. For instance, it could not write proofs itself, and the user chose one of three effort levels that set how far it continued on its own. The final design keeps the first rule, now enforced through the orchestrator's tool permissions, and drops the effort levels. Every session is autonomous, delegated regions and builds return a handle immediately and run in the background, and the orchestrator decides what to delegate next and when to collect each result.

\begin{figure}[t]
\centering
% Agent layers in autonomous mode. Drawn at true size for a 5.5in text block.
\providecommand{\figsub}{\scriptsize\sffamily\color{nMuted}}
\begin{tikzpicture}[
  font=\footnotesize\sffamily,
  >={Stealth[length=1.5mm]},
  node/.style={draw=nAccent!70, fill=white, rounded corners=2pt, line width=0.5pt,
    minimum height=7mm, align=center, text=nInk, inner sep=2pt},
  queued/.style={node, draw=nGray, dashed, text=nGray, fill=nPaper},
  prover/.style={draw=nAccent, fill=nAccent, text=white, rounded corners=1.5pt,
    minimum width=5.2mm, minimum height=6mm, inner sep=0pt, font=\scriptsize\sffamily\bfseries},
  link/.style={line width=0.6pt, draw=nAccent!70},
  qlink/.style={line width=0.5pt, draw=nGray, dashed},
  layer/.style={anchor=west, text width=3.9cm, align=left, inner sep=0pt},
]
% row guides
\foreach \y in {0,-1.45,-2.9,-4.35}
  \draw[nBorderLight, line width=0.4pt] (0,\y-0.72) -- (13.9,\y-0.72);

% layer descriptions (left column)
\node[layer] at (0,0)     {\textbf{Orchestrator}\\[-1pt]{\figsub a request, or a whole blueprint}\\[-3pt]{\figsub runs in the background}};
\node[layer] at (0,-1.45) {\textbf{Regional child}\\[-1pt]{\figsub one region, own Git worktree}\\[-3pt]{\figsub up to four run at once}};
\node[layer] at (0,-2.9)  {\textbf{Role agents}\\[-1pt]{\figsub long-lived writers and reviewers,}\\[-3pt]{\figsub explorer, golfer}};
\node[layer] at (0,-4.35) {\textbf{Provers}\\[-1pt]{\figsub one declaration each}\\[-3pt]{\figsub typically 3--6\,min, $\le$8 at once}};

% orchestrator
\node[node, minimum width=3.0cm, fill=nAccentSoft] (orch) at (9.5,0) {\textbf{orchestrator}\\[-2pt]{\figsub autonomous session}};

% regions: one active, the rest queued behind it
\node[node, minimum width=2.4cm, line width=0.8pt] (r1) at (6.4,-1.45) {\textbf{region 1}\\[-2pt]{\figsub own worktree}};
\node[node, minimum width=2.4cm] (r2) at (9.5,-1.45) {\textbf{region 2}\\[-2pt]{\figsub own worktree}};
\node[node, minimum width=2.4cm] (r3) at (12.6,-1.45) {\textbf{region 3}\\[-2pt]{\figsub own worktree}};
\draw[link] (orch.south) -- (9.5,-0.92);
\draw[link] (9.5,-0.92) -- (6.4,-0.92);
\draw[link, ->] (6.4,-0.92) -- (r1.north);
\draw[link] (9.5,-0.92) -- (12.6,-0.92);
\draw[link, ->] (9.5,-0.92) -- (r2.north);
\draw[link, ->] (12.6,-0.92) -- (r3.north);

% role agents under region 1
\node[node, minimum width=2.7cm] (wr) at (5.45,-2.9) {writer $\rightleftarrows$ reviewer\\[-2pt]{\figsub per-declaration review}};
\node[node, minimum width=1.5cm] (ex) at (7.75,-2.9) {explorer};
\node[node, minimum width=1.3cm] (go) at (9.3,-2.9) {golfer};
\draw[link] (r1.south) -- (6.4,-2.3);
\draw[link] (5.45,-2.3) -- (10.45,-2.3);
\foreach \n in {wr,ex,go} \draw[link, ->] (\n.north |- 0,-2.3) -- (\n.north);

% prover batch under the active region
\foreach \i in {1,...,8}
  \node[prover] (p\i) at ({4.6+0.75*(\i-1)},-4.35) {P\i};
\draw[link] (10.45,-2.3) -- (10.45,-3.75);
\draw[link] (p1.north |- 0,-3.75) -- (10.45,-3.75);
\foreach \i in {1,...,8}
  \draw[link, ->] (p\i.north |- 0,-3.75) -- (p\i.north);

\end{tikzpicture}
\caption{Agent layers in a session that delegates. The orchestrator delegates regions of the blueprint to child orchestrators, each in its own Git worktree; up to four run at once (one by default), and their patches are merged into the parent one at a time. Inside each region, specialists and a batch of provers do the work (shown for one region). Host-wide and per-user slot leases bound the total number of child processes.}
\label{fig:layers}
\end{figure}
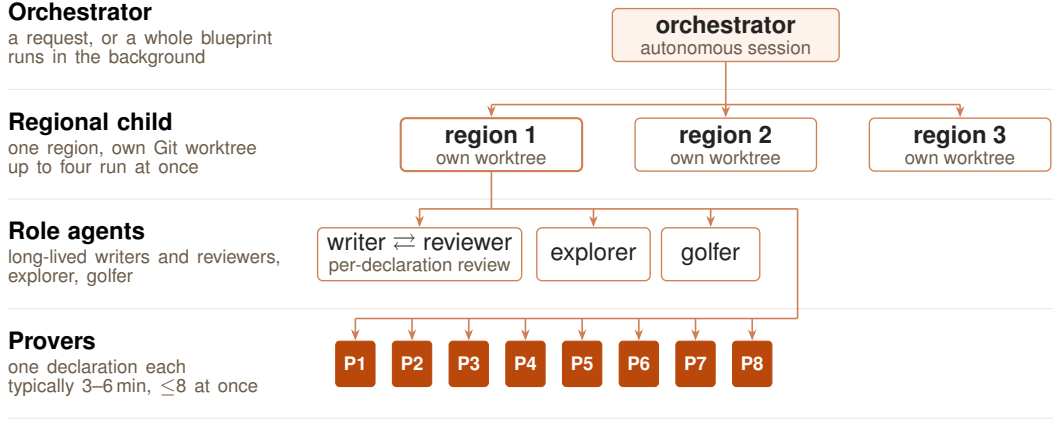

\paragraph{Delegation.} The orchestrator delegates every edit, down to a single lemma, as a region handled by a child orchestrator, shown in Figure~\ref{fig:layers}. A delegated task is recorded in the database and receives a detached worktree of the parent's clone, so a child's partial or failed work never touches the parent. Up to four children can run at once (one by default), and a new child starts as soon as another finishes; their patches are applied one at a time under a lock, conflicts in files that only list imports or \code{\textbackslash input} entries are resolved automatically, and other conflicts are kept as recovery bundles rather than failing the child. Delegation is one level deep, and children cannot delegate further.

\begin{figure}[!b]
\centering
% METR 50% time horizons (Time Horizon 1.1 data, values in minutes with 95% CIs),
% with the period in which Fuse was built shaded. Dates are decimal years of
% model release. Source: metr2026th11, metr2026timehorizons.
\begin{tikzpicture}
\begin{axis}[
  width=13.9cm, height=4.8cm,
  scale only axis=false,
  xmin=2025.0, xmax=2026.84,
  ymode=log, ymin=20, ymax=3700,
  xtick={2025.0,2025.25,2025.5,2025.75,2026.0,2026.25,2026.5,2026.75},
  xticklabels={Jan 2025,Apr,Jul,Oct,Jan 2026,Apr,Jul,Oct},
  ytick={30,60,120,240,480,960,1920},
  yticklabels={30 min,1 h,2 h,4 h,8 h,16 h,32 h},
  log ticks with fixed point,
  tick label style={font=\scriptsize\sffamily, text=nMuted},
  ylabel={50\% time horizon}, ylabel style={font=\scriptsize\sffamily, text=nMuted},
  axis line style={draw=nBorder}, tick style={draw=nBorder},
  ymajorgrids, grid style={draw=nBorderLight, line width=0.4pt},
  axis lines*=left,
  clip=false,
  legend style={font=\scriptsize\sffamily, draw=none, fill=none, at={(0.52,0.04)}, anchor=south, text=nInk, /tikz/column sep=10pt},
  legend columns=3,
  legend cell align=left,
]
% Fuse build period: 2026-03-10 to 2026-08-11
\fill[nAccentSoft] (axis cs:2026.19,20) rectangle (axis cs:2026.61,3700);
\node[font=\scriptsize\sffamily\bfseries, text=nAccent, anchor=north] at (axis cs:2026.43,3500) {Fuse built};
\node[font=\scriptsize\sffamily, text=nAccent, anchor=north] at (axis cs:2026.43,2350) {Mar--Aug 2026};
% METR's reliable-measurement ceiling
\draw[dashed, draw=nGray, line width=0.5pt] (axis cs:2025.0,960) -- (axis cs:2026.84,960);
\node[font=\scriptsize\sffamily, text=nMuted, anchor=south west] at (axis cs:2025.0,960) {16 h: limit of METR's current task suite};

% Least-squares exponential fit to the frontier (models that set a new record at release):
% 3.7 Sonnet, o3, GPT-5, Gemini 3 Pro, Opus 4.5, GPT-5.2, Opus 4.6, Mythos Preview.
% horizon = 46.99 min * 2^(3.2681 * (year - 2025)), i.e. doubling every 112 days.
\addplot[domain=2025.0:2026.27, samples=2, draw=nInk!55, line width=0.8pt, forget plot] {46.99*2^(3.2681*(x-2025))};
\addplot[domain=2026.27:2026.84, samples=2, draw=nInk!55, line width=0.8pt, dashed, forget plot] {46.99*2^(3.2681*(x-2025))};

% Other frontier models: open gray circles
\addplot[only marks, mark=o, mark size=2.2pt, mark options={draw=nGray, fill=white, line width=0.6pt},
  error bars/.cd, y dir=both, y explicit, error mark=none, error bar style={draw=nGray!55, line width=0.5pt}]
  coordinates {
    (2025.29,119.7) += (0,71.2) -= (0,45.1)   % o3
    (2025.60,203.0) += (0,202.6) -= (0,90.4)  % GPT-5
    (2025.88,224.3) += (0,154.9) -= (0,84.7)  % Gemini 3 Pro
    (2025.885,223.7) += (0,172.5) -= (0,89.4) % GPT-5.1-Codex-Max
    (2025.94,352.2) += (0,463.0) -= (0,154.1) % GPT-5.2
    (2026.10,349.5) += (0,466.9) -= (0,154.6) % GPT-5.3-Codex
    (2026.13,384.1) += (0,310.7) -= (0,150.6) % Gemini 3.1 Pro
    (2026.17,341.7) += (0,427.1) -= (0,155.1) % GPT-5.4
  };
\addlegendentry{Other frontier models}
% Claude models: filled accent circles
\addplot[only marks, mark=*, mark size=2.4pt, mark options={draw=white, fill=nAccent, line width=0.6pt},
  error bars/.cd, y dir=both, y explicit, error mark=none, error bar style={draw=nAccent!50, line width=0.6pt}]
  coordinates {
    (2025.15,60.4) += (0,43.8) -= (0,27.4)      % Claude 3.7 Sonnet
    (2025.39,100.4) += (0,63.1) -= (0,40.4)     % Claude Opus 4
    (2025.90,293.0) += (0,330.7) -= (0,131.3)   % Claude Opus 4.5
    (2026.10,718.8) += (0,2915.0) -= (0,402.1)  % Claude Opus 4.6
    (2026.27,1044.8) += (0,2259.5) -= (0,535.9) % Claude Mythos Preview (early)
  };
\addlegendentry{Claude models}
\addlegendimage{draw=nInk!55, line width=0.8pt}
\addlegendentry{frontier trend, doubling every 112 days}
% direct labels for Claude points
\begin{scope}[every node/.style={font=\scriptsize\sffamily, text=nInk}]
\node[anchor=west]  at (axis cs:2025.165,60.4)  {3.7 Sonnet, 1 h};
\node[anchor=west]  at (axis cs:2025.405,100.4) {Opus 4, 1.7 h};
\node[anchor=east]  at (axis cs:2025.885,293.0) {Opus 4.5, 4.9 h};
\node[anchor=east]  at (axis cs:2026.085,718.8) {Opus 4.6, 12 h};
\node[anchor=north west, inner sep=1pt]  at (axis cs:2026.285,900) {Mythos Preview, 17 h};
\end{scope}
\end{axis}
\end{tikzpicture}
\caption{METR 50\% time horizons of frontier models by release date, with 95\% confidence intervals~\cite{metr2026th11,metr2026timehorizons}: the length of task, in skilled-human time, that a model completes half the time. The line is an exponential fit to the models that set a new record when released, doubling every 112 days (METR estimates about 129 days for all models since 2023), dashed where extrapolated. The shaded band is the period in which Fuse was built.}
\label{fig:horizons}
\end{figure}
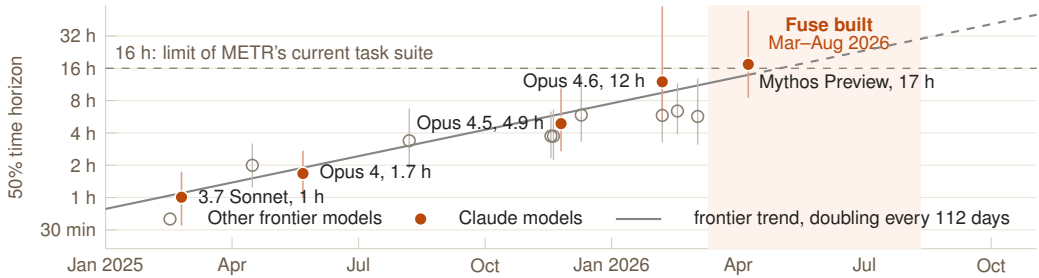

\paragraph{Why so many layers.} Decomposition provided parallelism, routed proof search to cheaper models, allowed independent review, and contained failures in worktrees. It also compensated for agents that, in early 2026, did not reliably finish long tasks. The effort levels re-prompted an agent that stopped early, and checkpoints let interrupted writer/reviewer loops resume. As agent time horizons rose (Figure~\ref{fig:horizons}), the mechanisms that existed only to keep tasks short were the first ones Fuse removed. Effort levels gave way to sessions that are always autonomous, and writer/reviewer loops enforced in code gave way to standing specialists. Apart from the prover study and the golfer's heartbeat reductions, none of these design changes was measured.

%=====================================================================
\section{Infrastructure and operations}
\label{sec:infra}

\paragraph{Lean builds and language servers.} Building a workspace installs the pinned toolchain, restores dependencies from a shared cache keyed by package, revision, and toolchain (copied into each clone as copy-on-write reflinks), fetches the Mathlib build cache, and runs \code{lake build}. Build errors are stored in a snapshot keyed by file hash, which lets a file open with its diagnostics before Lean finishes elaborating, and blocks a commit only when a changed \code{.lean} file has a known error. Each workspace clone gets one Lean language server, shared by the agents and the Lean view in the browser, so both see the same proof state; child orchestrators check their worktrees with separate \code{lake env lean} runs.

\paragraph{Git and collaboration.} The backend keeps one sparse clone per workspace, evicting clean clones when disk runs low. Commits happen after each agent turn, on a manual commit or save, and when the last editor disconnects, and are pushed with \code{--force-with-lease}; a rejected push is rebased and revalidated, and a small agent attempts any rebase conflict. Merging another branch is a single background transaction run by a bounded merger agent that ends in a commit or a rollback. The blueprint editor uses Yjs with the authoritative document held on the server, so people and agents can edit the same blueprint at once.

\paragraph{Metering and budgets.} Every model call is metered once, including side channels such as OCR and title generation, and a watchdog stops a user's sessions when their daily or weekly cap is reached. Users may instead link their own Claude subscription, in which case the spend of every Claude agent on it is excluded from Fuse's caps.

\paragraph{Deployment.} Long autonomous sessions conflict with frequent deploys. Fuse runs each deployed commit as its own release; session identifiers carry a route tag, and nginx sends a session's traffic to the release that owns it while new sessions go to the current one. Deploys that change the database schema instead pause running sessions, record how to resume them, and restart them on the new release with a resumption prompt.

%=====================================================================
\section{Evaluation and deployment experience}
\label{sec:eval}

\subsection{Prover selection and cost study}
\label{sec:provers}

Proof search dominated cost, so we compared prover models in the production prover configuration (Table~\ref{tab:bench}). The 52 lemmas are every lemma and theorem in four fully proved blueprint developments (Catalan generating functions, the finite Ramsey theorem, the fundamental group of the circle, and connectivity of Cayley graphs), each restated with a \code{sorry} in a fresh directory containing only the proved material before it. The agents' shell could still reach the original source on disk, but the three to eight solutions per model identical to an original proof are all one- to three-line proofs, most found by several models. We approximate difficulty by original proof length, with easy at most 4 lines (16 lemmas), medium at most 15 (23), and hard longer (13). Each model made one attempt per lemma inside Claude Code with the prover's tools, limited to 80 turns and 30 minutes. A proof counts only if \code{leanprover/comparator} replays it in the kernel using only \code{propext}, \code{Quot.sound}, and \code{Classical.choice}. Cost per solved lemma divides all spending, including failed attempts, by the number solved. We set aside six runs made while the Mathlib search service was down and reran only Claude Opus~4.8. Network timeouts (three for DeepSeek V4 Flash) and an unresponsive provider (three hard lemmas for DeepSeek V4 Pro) count as failures. Appendix~\ref{app:provers} gives the full setup, prices, and totals.

\begin{table}[t]
\centering\small\tablestyle
\caption{Prover models on 52 lemmas, one attempt each, run inside Claude Code with the production prover tools in May--July 2026. Cost per solved lemma includes failed attempts and uses pinned provider prices.}
\label{tab:bench}
\begin{tabular}{@{}l@{\hspace{6pt}}r@{\hspace{5pt}}l@{\hspace{10pt}}ccc@{\hspace{10pt}}r@{\hspace{5pt}}l@{}}
\toprule
\thead{Model} & \multicolumn{2}{l}{\thead{Solved}} & \thead{Easy} & \thead{Medium} & \thead{Hard} & \multicolumn{2}{l}{\thead{\$ per solved lemma}} \\
 & & & {\scriptsize\color{nMuted}of 16} & {\scriptsize\color{nMuted}of 23} & {\scriptsize\color{nMuted}of 13} & & \\
\midrule
Doubao Seed 2.1 Turbo & 52/52 & \solvedbar{52}{52} & 16 & 23 & 13 & 0.080 & \costbar{0.08} \\
\rowrule
Claude Sonnet 4.6 & 51/52 & \solvedbar{51}{52} & 16 & 23 & 12 & 0.421 & \costbar{0.421} \\
\rowrule
Doubao Seed 2.1 Pro & 51/52 & \solvedbar{51}{52} & 16 & 23 & 12 & 0.162 & \costbar{0.162} \\
\rowrule
Claude Opus 4.8 & 49/52 & \solvedbar{49}{52} & 15 & 22 & 12 & 0.526 & \costbar{0.526} \\
\rowrule
Kimi K2.6 & 49/52 & \solvedbar{49}{52} & 16 & 22 & 11 & 0.398 & \costbar{0.398} \\
\rowrule
DeepSeek V4 Flash & 47/52 & \solvedbar{47}{52} & 14 & 23 & 10 & 0.041 & \costbar{0.041} \\
\rowrule
DeepSeek V4 Pro & 45/52 & \solvedbar{45}{52} & 15 & 20 & 10 & 0.218 & \costbar{0.218} \\
\bottomrule
\end{tabular}
\end{table}

All seven models solved between 45 and 52 of the 52 lemmas, so this set does not separate them reliably; a single run cannot order 49, 51, and 52. Cost varied by more than an order of magnitude at similar accuracy, which is why proof search goes to inexpensive models while a frontier model orchestrates.

\subsection{Kakeya in dimension three}
\label{sec:kakeya}

\paragraph{Scope.} The formalization proves that every compact set in $\mathbb{R}^3$ containing a unit segment in every direction has Hausdorff dimension three, following the streamlined argument of Guth, Wang, and Zahl~\cite{guth2026streamlined}. The main theorem takes one input as an explicit hypothesis, \code{StickyFrostmanHypothesis}, the project's formulation of their sticky Frostman estimate (Theorem~7.3(A)), so its type is
\begin{center}
\code{KakeyaDimensionThree : StickyFrostmanHypothesis~$\to$ KakeyaSetConjecture~3}.
\end{center}
The library declares no axioms; the release's verification script rebuilds it and checks that the main theorem depends only on Lean's three standard axioms (\code{propext}, \code{Classical.choice}, and \code{Quot.sound}). The released repository discharges the hypothesis with an independent formalization by a Nankai University and ByteDance Seed team, giving the unconditional theorem \code{Unconditional.KakeyaDimensionThree}, of type \code{KakeyaSetConjecture~3}~\cite{chen2026sticky,numina2026kakeya3d}. The pinned release (commit \code{a26427e}, Lean \code{v4.32.0-rc1}) has about 1{,}100 Lean files and 650k lines.

\paragraph{The Fuse phase.} From May to early September, the Kakeya team worked directly with the Fuse developers, and much of Fuse's design was shaped by what they needed. Most of the project's definitions and main theorem statements were formalized in this phase, with contributors writing and reviewing statements in the blueprint and in Lean while Fuse agents proved and golfed them. Problems the team found in review, such as agents strengthening statements they could not prove or re-proving lemmas instead of reusing them, led directly to strengthening Fuse's reviewer checks and creating the explorer agent. The golfer also reduced the elaboration cost of one central file from about 220M to 85M heartbeats (Lean's count of elaboration work, not wall-clock time), a reduction of about 61\%.

\paragraph{The final assembly.} The last two weeks, from August~26 to September~7, were done outside Fuse. One contributor directed Claude Code with the newly released Claude Fable~5.1, running agents in parallel Git worktrees with a reviewer agent. Fable~5.1 found statement errors that earlier review had missed. For example, the Lean statement of Proposition~6.6 of the paper had required the convex hull of a family of tubes to equal a plank, which can never hold; the agent proved it vacuously, and the statement was repaired, proved, and certified to have satisfiable hypotheses.

\paragraph{Human-paced and agent-paced work.} Fuse was built for one person formalizing mathematics with an AI and reviewing the work as it proceeds, while some recent systems formalize almost entirely autonomously~\cite{rammal2026formalizing,gloeckle2026textbook}. The final assembly was closer to the second mode, adding about 592k lines of Lean in 12 days. Excluding the roughly 35\% copied from other branches leaves at most 288k new lines, or 25k per day. By comparison, one of us used Fuse on the LeanEval benchmark, one session at a time, and its agents wrote about 69k lines across four workspaces, roughly 10k per day per session. The final assembly therefore wrote two to five times as much per day.

%=====================================================================
\section{Related work}
\label{sec:related}
\paragraph{Blueprint-driven formalization.} Formalizing against an informal blueprint dates to Flyspeck~\cite{hales2012densesphere,hales2017flyspeck}, whose blueprint and formal proof were developed together and cross-linked by identifiers. Lean projects adopted the idea through leanblueprint~\cite{massot2020leanblueprint}: the Liquid Tensor Experiment~\cite{commelin2022lte}, the FLT project~\cite{buzzard2024flt}, the Carleson project~\cite{becker2024carleson}, and the Equational Theories Project~\cite{bolan2025etp}, which combined human, automated, and Lean effort at scale. Fuse keeps this artifact and workflow, assigns drafting, formalization, and proof search to agents, and leaves the statements to people. LeanArchitect~\cite{zhu2026leanarchitect} goes in the other direction, generating blueprints from annotated Lean.

\paragraph{Agentic autoformalization systems.} Several systems from 2026 share Fuse's structure of a blueprint, specialized agents, and review. LeanMarathon~\cite{zhang2026leanmarathon} maintains an evolving blueprint with construct, audit, prove, and repair agents and adversarial fidelity review; FormalFlow~\cite{lu2026mipstar} runs nested plan--prove--review loops under human supervision; Goedel-Architect~\cite{chung2026goedelarchitect} refines a blueprint graph and closes lemmas in parallel with an open-weight prover, close to Fuse's prover batches; MathCoPilot~\cite{zhang2026mathcopilot}, TNLean~\cite{lu2026tensor}, and Ax-Prover~\cite{breen2025axprover} keep a human in the loop to varying degrees. Math Inc's Gauss completed the Strong Prime Number Theorem project with targeted human scaffolding~\cite{mathinc2025strongpnt}. At the other extreme, Meta's AutoformBot/Atlas runs thousands of autonomous agents with shared version control over whole textbooks, producing more than 45{,}000 declarations~\cite{rammal2026formalizing,gloeckle2026textbook}. Fuse differs mainly in keeping the blueprint as an ordinary leanblueprint document in the user's repository, in its explicit golfing and style stages, and in splitting a frontier coordinator from inexpensive provers.

\paragraph{Provers.} Trained provers range from whole-proof models (DeepSeek-Prover~\cite{xin2024deepseekprover,ren2025deepseekprover2}, Kimina-Prover~\cite{wang2025kimina}, Goedel-Prover~\cite{lin2025goedel,lin2025goedel2}, Seed-Prover~\cite{chen2025seedprover}) to step-level search~\cite{xin2025bfsprover,wu2024internlm25stepprover} and RL-trained systems such as AlphaProof~\cite{hubert2025alphaproof}, Aristotle~\cite{achim2025aristotle}, and Seed-Prover~1.5~\cite{chen2025seedprover15}. Agentic frameworks descending from Draft-Sketch-Prove~\cite{jiang2022dsp} orchestrate general-purpose models, sometimes alongside specialized provers; they include APOLLO~\cite{ospanov2025apollo}, Delta-Prover~\cite{zhou2025deltaprover}, and Hilbert~\cite{varambally2025hilbert}; a minimal agent baseline is competitive with them~\cite{requena2026minimal}. Competition benchmarks are now saturated: several systems report all 672 PutnamBench~\cite{tsoukalas2024putnambench} problems~\cite{putnambench_leaderboard}, and miniF2F~\cite{zheng2021minif2f} is effectively solved~\cite{chung2026goedelarchitect,varambally2025hilbert}. Fuse treats such models as interchangeable workers rather than competing with them. Its Lean tooling builds on LeanDojo~\cite{yang2023leandojo}, Lean Copilot~\cite{song2024leancopilot}, Pantograph~\cite{aniva2024pantograph}, LeanExplore~\cite{asher2025leanexplore}, and Loogle~\cite{breitner2023loogle}.

\paragraph{Statement faithfulness.} Type-checking overstates faithfulness. A revision of miniF2F found mismatches in more than half of its formal statements~\cite{ospanov2025minif2frevisited}; on 227 university-level textbook statements, the gap between formalizations that compile and those accepted as faithful reaches 29.5 points for one agent, and the acceptance check, compilation combined with agreement between two LLM judges, matches the human majority in 91.5\% of audited cases~\cite{zhang2026beyondcompilation}. Counterexample search catches some misformalizations automatically~\cite{shen2026mechgeo,firsching2026formalconjectures}, which is why Fuse's blueprint reviewer looks for counterexamples first. Within hours of release, experts reviewing the unreviewed Atlas library found theorems whose content was assumed as structure fields, trivialized definitions, and a tautological reduction rated 5/5 by the system's LLM judge~\cite{zulip2026atlasthread}. Klowden and Tao note that formal verification does not rule out errors in translating the intended statement, so human review of statements remains necessary~\cite{klowden2026methods}, and the Gemini Erdős study found 68.5\% of 200 candidate solutions fundamentally flawed~\cite{feng2026aletheiaerdos}.

\paragraph{Agent harnesses and multi-agent systems.} The bitter lesson~\cite{sutton2019bitter} predicts that hand-built structure loses to general methods that scale with compute. For coding agents, simple scaffolds match elaborate ones~\cite{xia2024agentless,minisweagent}, and commercial harnesses do not beat METR's plain scaffolds~\cite{jurkovic2026claudecodecodex}. Anthropic reports that harness components ``can quickly go stale as models improve'', citing context resets and sprint-based task decomposition dropped with newer models~\cite{rajasekaran2026harnessdesign}. Hand-designed role-based multi-agent systems~\cite{wu2023autogen,hong2023metagpt,qian2024chatdev} often yield small or task-dependent gains~\cite{cemri2025why,kim2025scaling}; orchestrator--worker systems with a strong lead model show clear gains on breadth-heavy research~\cite{hadfield2025multiagent}. Models are now trained for multi-agent operation directly: Kimi~K2.5 learns with reinforcement learning to run subagents in parallel~\cite{kimi2026k25}, and Claude Code subagents can now spawn their own, up to a configurable depth~\cite{anthropic2026subagents}. Fuse instead starts every child through its own MCP tools, so that each inherits Fuse's permission checks.

%=====================================================================
\section{Lessons and outlook}
\label{sec:lessons}

Formalization is becoming easier to produce than to absorb into a useful, shared mathematical resource.

\paragraph{Blueprints are now a bottleneck.} Fuse was built around the blueprint as the plan agents follow and the place people review it. With current models, a separately maintained blueprint adds work, falls out of date, and constrains how agents decompose the problem. Formalizing statement by statement, with each blueprint entry drafted, reviewed, formalized, and proved in turn, is especially slow, and it forces the system to track and synchronize state for every declaration. We expect a graph of Lean statements, as in Prove2Me~\cite{chen2026prove2me}, to be more efficient. Agents choose what to attempt next from the graph and reuse existing results, and the decomposition develops and changes as they write Lean, without a separate document to maintain.

\paragraph{People should set standards, not review every step.} Fuse was designed around a person checking each step. Once models wrote compiling Lean far faster than people could read it, that loop became the bottleneck: even an engineer reading ten thousand lines a day would need two months for a project the size of Kakeya, and its final assembly ran at two to five times that rate (Section~\ref{sec:kakeya}). The statement errors in Section~\ref{sec:using} were all found by people, so this experience does not support removing them from the loop. For now, routine checkpoints on intermediate lemmas can go, and human review should concentrate on definitions, main theorem statements, and accepting a release. People should also write the standards agents work to: what needs to be formalized, and what a good formalization looks like, from how definitions are chosen to how results are stated and organized.

\paragraph{Architecture and search are more important.} As agents get better at writing correct, compiling Lean, the hard problem shifts from the quality of individual proofs to the architecture of the library as a whole. Fuse's golfer and explorer worked at the level of single proofs and lemmas, but at scale the questions are which definitions to build on, how a development fits with what already exists, and whether its API is one that later projects can use. In addition, people and agents need to be able to access and search this API, finding existing results by informal description or by type and recognizing equivalent formulations~\cite{asher2025leanexplore,breitner2023loogle}, so that new work extends the library instead of duplicating it.

\paragraph{Progress now depends on infrastructure and institutions.} Progress may now be limited less by proof generation than by access to papers and compute and by the work of building, verifying, and maintaining formal mathematics across toolchain changes. This calls for a coordinated initiative to formalize mathematics comprehensively: a common undertaking with sustained resources and a shared destination for the results, in the spirit of the proposed Mathematics Autoformalization Project~\cite{lichtman2026map}. Together, these point toward less prescribed scaffolding, more autonomous production, and more investment in the infrastructure and institutions that make formalized mathematics accessible, trustworthy, and cumulative.

%=====================================================================
\section{Conclusion}
\label{sec:conclusion}

Fuse connected human statement review, agent proof search, persistent project state, and Lean verification in one environment, and over 2026 it moved from a prescribed pipeline toward an orchestrator that chooses its own dispatch. It supported the Kakeya formalization from May until its final assembly, and its prover study showed that proof search can go to inexpensive models at similar accuracy. The parts of Fuse that lasted were the ones that did not depend on a model: the project state, the Lean tooling, and verification.

{\small
\bibliographystyle{plainnat}
\bibliography{references}
}

\clearpage
\appendix
%=====================================================================
\section{Prover study data}
\label{app:provers}

\paragraph{Setup.} The 52 lemmas come from four developments in one Lean
project on Lean \code{v4.30.0-rc2} and Mathlib \code{7ea6142b3f}. For each
lemma, a fresh directory holds a \code{Machinery} module with every
declaration before the target, proved, and a \code{Challenge} module with the
target stated with \code{sorry}; the agent fills a copy of \code{Challenge}.
Every model ran through the Claude Code agent SDK with the production prover
prompt and the tools Read, Edit, Write, Bash, Grep, Glob, the Lean language
server, and lean-explore, with at most 80 turns and 30 minutes per attempt,
four attempts at a time, and one attempt per lemma. Non-Anthropic models were
reached through Anthropic-compatible endpoints, pinned to the providers in
Table~\ref{tab:prices}. The harness version was not recorded. A solution
counts only if \code{leanprover/comparator} accepts it with axioms limited to
\code{propext}, \code{Quot.sound}, and \code{Classical.choice}.

\begin{table}[H]
\centering\small\tablestyle
\caption{Prices used for cost, in USD per million tokens. Volcengine Ark
prices are converted from RMB at about 7.2 CNY per USD. Cache writes were free
on the WandB and Alibaba endpoints.}
\label{tab:prices}
\begin{tabular}{@{}llrrr@{}}
\toprule
\thead{Model} & \thead{Provider} & \thead{Input} & \thead{Output} & \thead{Cache read} \\
\midrule
Doubao Seed 2.1 Turbo & Volcengine Ark & 0.417 & 2.083 & 0.083 \\
\rowrule
Claude Sonnet 4.6 & Anthropic & 3.00 & 15.00 & 0.30 \\
\rowrule
Doubao Seed 2.1 Pro & Volcengine Ark & 0.833 & 4.167 & 0.167 \\
\rowrule
Claude Opus 4.8 & Anthropic & 5.00 & 25.00 & 0.50 \\
\rowrule
Kimi K2.6 & WandB & 0.95 & 4.00 & 0.16 \\
\rowrule
DeepSeek V4 Flash & Alibaba & 0.134 & 0.268 & 0.0268 \\
\rowrule
DeepSeek V4 Pro & Alibaba & 1.608 & 3.216 & 0.134 \\
\bottomrule
\end{tabular}
\end{table}

\begin{table}[H]
\centering\small\tablestyle
\caption{Totals per model over the 52 attempts.}
\label{tab:prover-totals}
\begin{tabular}{@{}lrrrr@{}}
\toprule
\thead{Model} & \thead{Solved} & \thead{Total \$} & \thead{Mean turns} & \thead{Mean wall time (s)} \\
\midrule
Doubao Seed 2.1 Turbo & 52/52 & 4.17 & 20.0 & 242 \\
\rowrule
Claude Sonnet 4.6 & 51/52 & 21.45 & 21.2 & 249 \\
\rowrule
Doubao Seed 2.1 Pro & 51/52 & 8.24 & 17.8 & 175 \\
\rowrule
Claude Opus 4.8 & 49/52 & 25.80 & 24.1 & 172 \\
\rowrule
Kimi K2.6 & 49/52 & 19.51 & 31.7 & 275 \\
\rowrule
DeepSeek V4 Flash & 47/52 & 1.91 & 25.1 & 383 \\
\rowrule
DeepSeek V4 Pro & 45/52 & 9.80 & 23.0 & 431 \\
\bottomrule
\end{tabular}
\end{table}

\clearpage
%=====================================================================
% Interface gallery: screenshots of Fuse Desktop on the Kakeya formalization.

\section{Interface gallery}
\label{app:gallery}

This appendix shows the workspace views described in Section~\ref{sec:using}, taken from Fuse Desktop, the open-source local version, on the Kakeya formalization. In each view the chat with the orchestrator stays open on the left.

\begin{figure}[!htbp]
\centering
\fusescreenshot{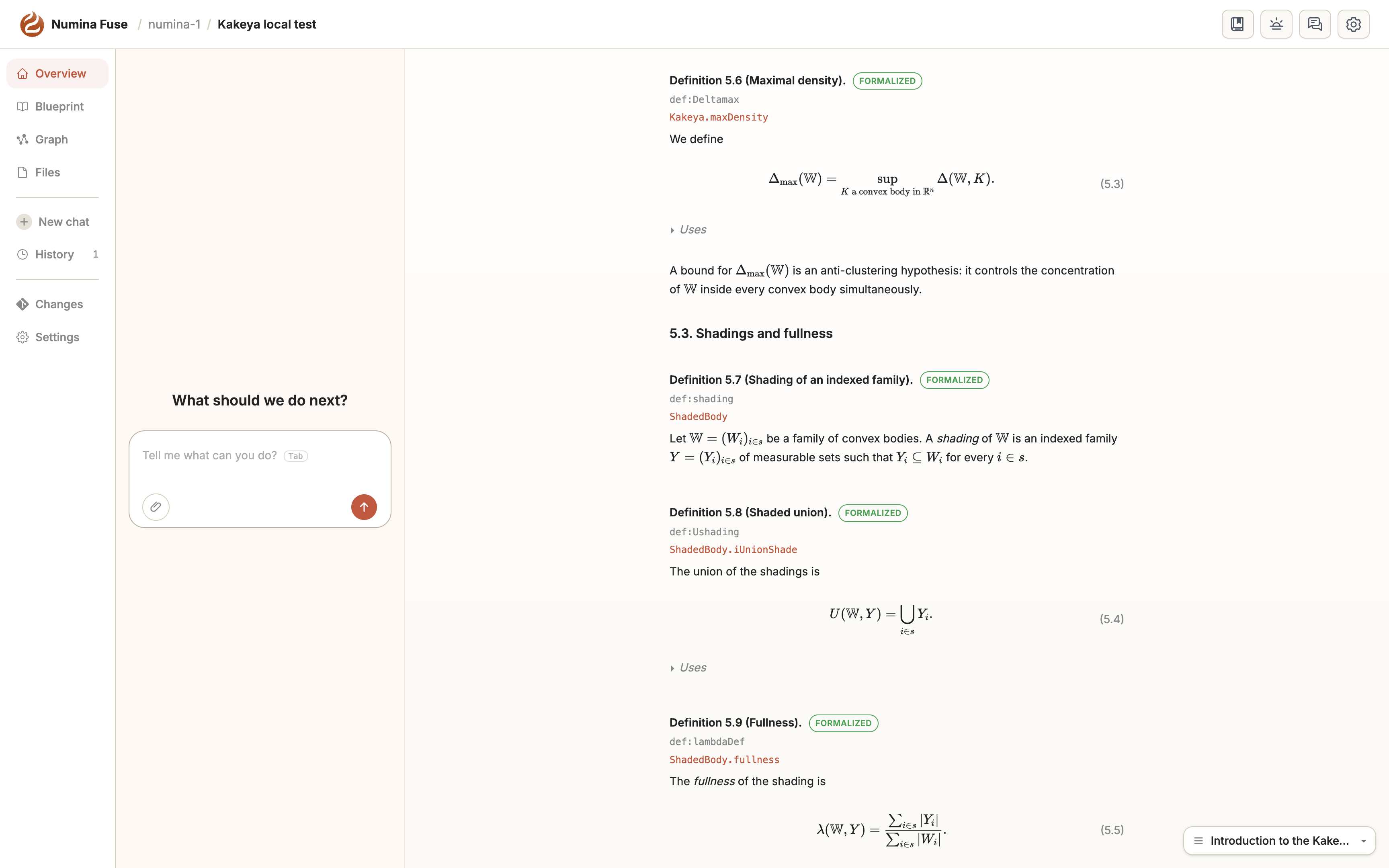}
\caption{The overview in Fuse Desktop, showing the blueprint as rendered text. Each entry carries its number, label, linked Lean declaration, and formalization status.}
\label{fig:gallery-overview}
\end{figure}

\begin{figure}[!htbp]
\centering
\fusescreenshot{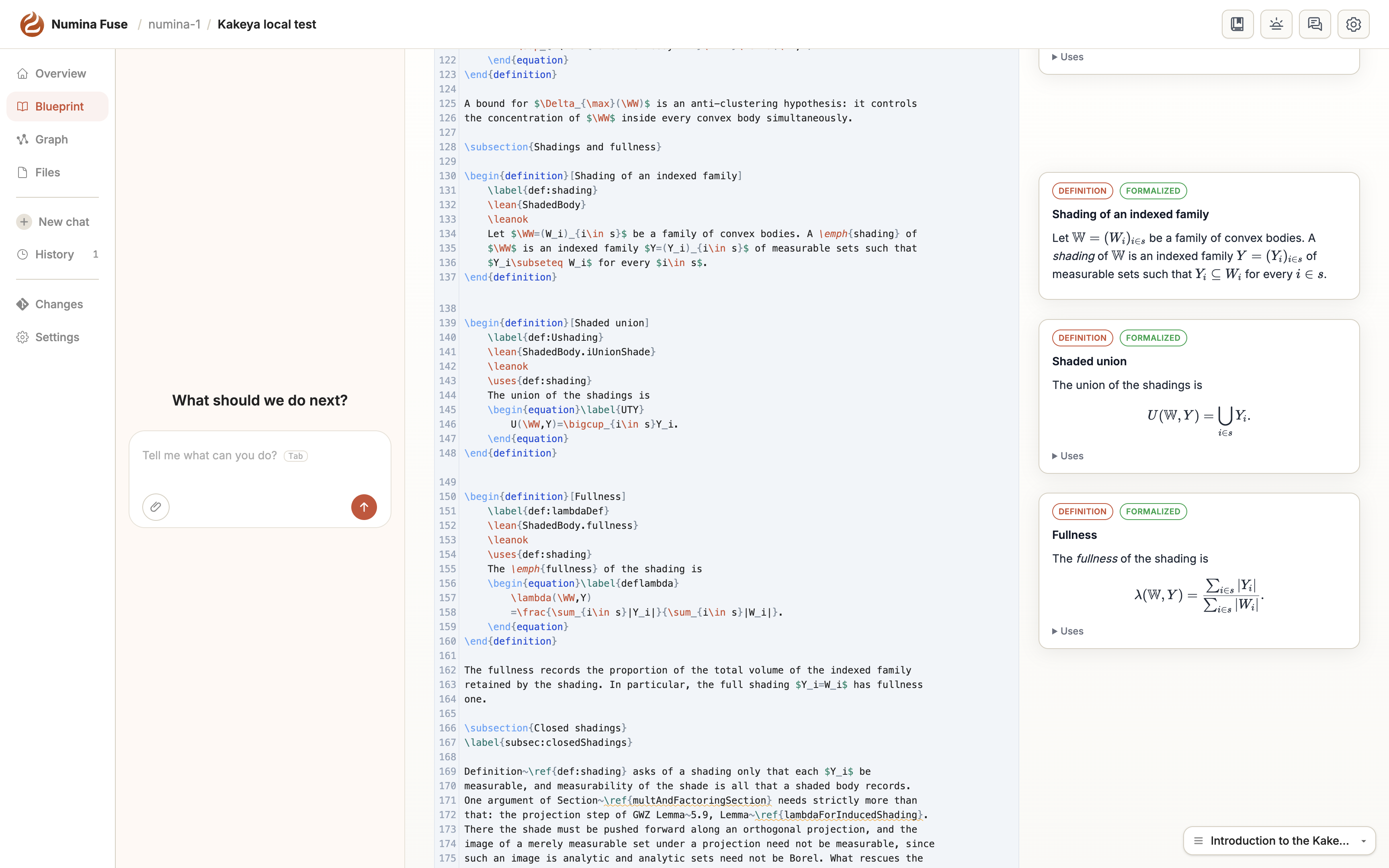}
\caption{The blueprint view in Fuse Desktop. The \LaTeX{} source, with its \code{\textbackslash lean} and \code{\textbackslash leanok} markers, sits beside a rendered card for each entry showing its kind and status.}
\label{fig:gallery-blueprint}
\end{figure}

\begin{figure}[!htbp]
\centering
\fusescreenshot{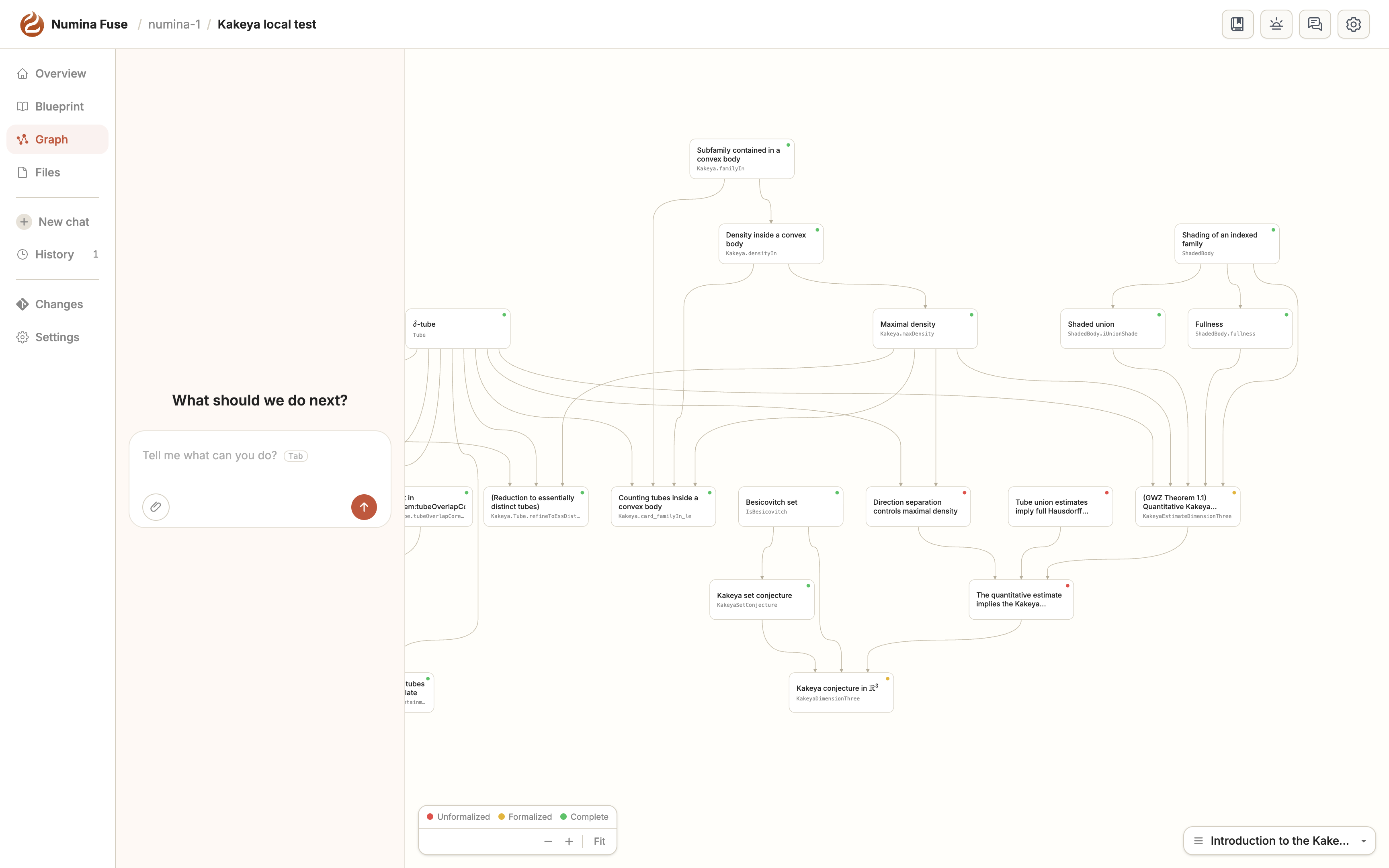}
\caption{The dependency graph in Fuse Desktop, colored by declaration status (unformalized, formalized, complete), with each node showing its linked Lean declaration.}
\label{fig:gallery-graph}
\end{figure}

\begin{figure}[!htbp]
\centering
\fusescreenshot{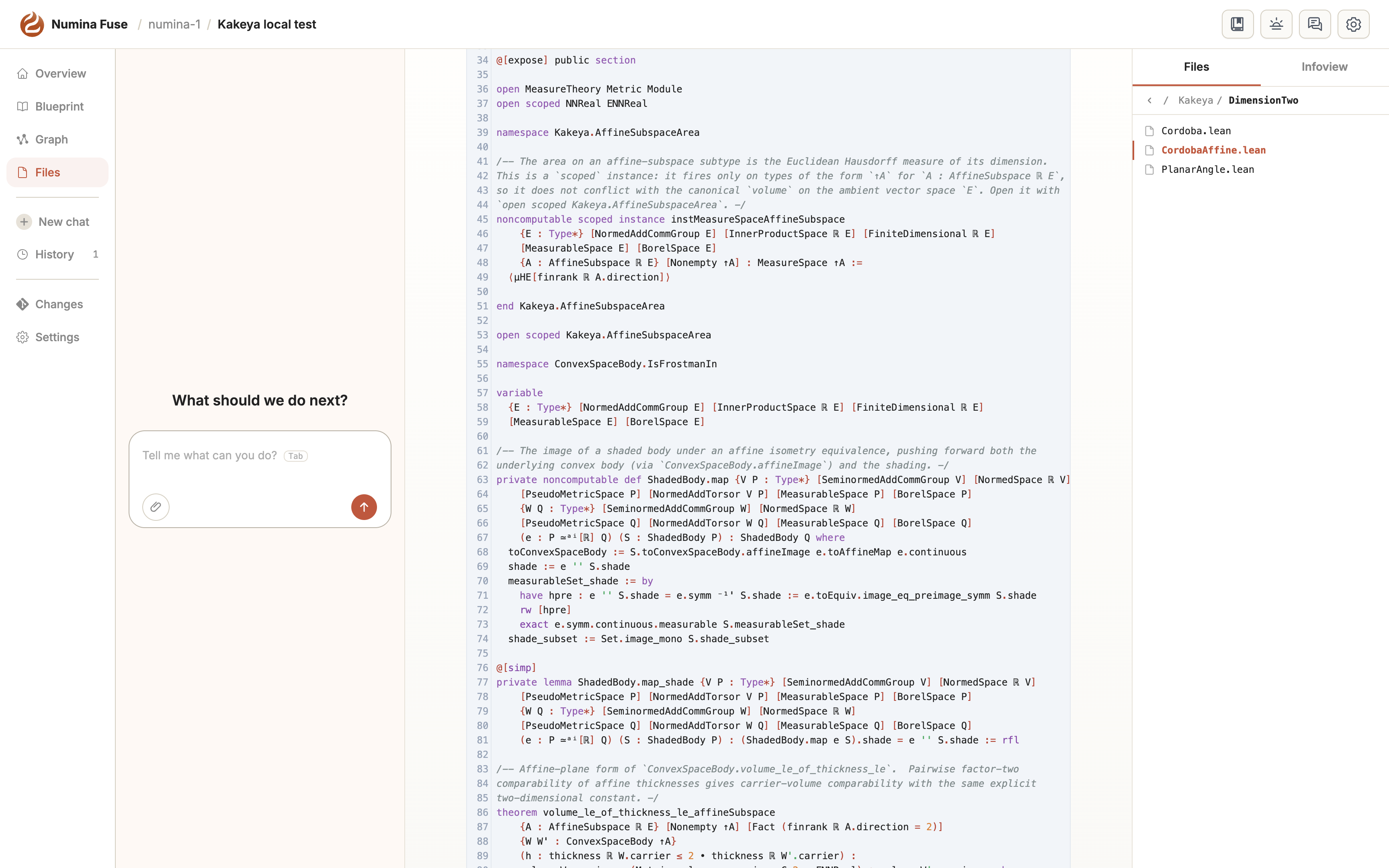}
\caption{The Lean view in Fuse Desktop, with the project's file browser and an infoview tab for the goal state from the language server.}
\label{fig:gallery-lean}
\end{figure}

\end{document}